\documentclass[preprint,12pt,sort&compress]{elsarticle}

\usepackage{textcomp,booktabs}
\usepackage[usenames,dvipsnames]{color}
\usepackage{colortbl}
\definecolor{mygray}{gray}{.9}
\definecolor{mypink}{rgb}{.99,.91,.95}
\definecolor{mycyan}{cmyk}{.3,0,0,0}
\usepackage{amssymb}
\usepackage{graphicx}
\usepackage{booktabs}
\usepackage{tabularx}
\usepackage{array}
\usepackage{lscape}
\usepackage{longtable}
\usepackage{multirow}
\usepackage{amsmath}
\usepackage{colortbl}
\usepackage[figuresright]{rotating}
\usepackage{epsfig}
\usepackage{epsf}
\usepackage{subfigure}
\usepackage{slashbox}

\newcommand{\PreserveBackslash}[1]{\let\temp=\\#1\let\\=\temp}
\newcolumntype{C}[1]{>{\PreserveBackslash\centering}p{#1}}
\newcolumntype{R}[1]{>{\PreserveBackslash\raggedleft}p{#1}}
\newcolumntype{L}[1]{>{\PreserveBackslash\raggedright}p{#1}}
\newtheorem{definition}{Definition}[section]
\newtheorem{example}{Example}[section]

\newtheorem{note}{Note}[section]
\journal{} 

\begin{document}

\begin{frontmatter}



\title{A new belief Markov chain model and its application in inventory prediction}


\author[address1]{Zichang He}
\author[address1]{Wen Jiang\corref{label1}}
\address[address1]{School of Electronics and Information, Northwestern Polytechnical University, Xi'an, Shaanxi, 710072, China}
\cortext[label1]{Corresponding author at: School of Electronics and Information, Northwestern  Polytechnical University, Xi'an, Shaanxi 710072, China. Tel: (86-29)88431267. E-mail address: jiangwen@nwpu.edu.cn, jiangwenpaper@hotmail.com}
\begin{abstract}
Markov chain model is widely applied in many fields, especially the field of prediction. The classical Discrete-time Markov chain(DTMC) is a widely used method for prediction. However, the classical DTMC model has some limitation when the system is complex with uncertain information or state space is not discrete. To address it, a new belief Markov chain model is proposed by combining Dempster-Shafer evidence theory with Markov chain. In our model, the uncertain data is allowed to be handle in the form of interval number and the basic probability assignment(BPA) is generated based on the distance between interval numbers. The new belief Markov chain model overcomes the shortcomings of classical Markov chain and has an efficient ability in dealing with uncertain information. Moreover, an example of inventory prediction and the comparison between our model and classical DTMC model can show the effectiveness and rationality of our proposed model.
\end{abstract}

\end{frontmatter}
\section{Introduction}\label{Introduction}
A Markov process can be used to model a random system that changes states according to a transition rule that only depends on the current state. And the Markov property is the most important property during Markov process, namely the conditional probability distribution of the future states of the process depends only upon the present state\cite{Kemeny1960Finite,Fu1994Distribution,Komorowski2010ON}. Markov chain is a stochastic process with Markov property\cite{Darling1953The}. It is widely used in many applications\cite{Annett2010Twelve,Alagoz2010Markov,Farahat2010Markov,Navaei2010MARKOV}, especially in the prediction\cite{Lu2014Application,Samet2014Enhancement}, such as rainfall prediction\cite{fraedrich1983single,Liu2011Study}, economy prediction\cite{jin2015phev,mar2010probabilistic} and so on.

Makov chain is a powerful tool to study sequential data. It provides a bunch of models to analyze time series data and predict the variation tendencies of random processes, including discrete-time Markov chain(DTMC)\cite{privault2013discrete}, continuous-time Markov chain\cite{suchard2001bayesian,aziz2000model}, hidden Markov chain\cite{krogh2001predicting,rabiner1986introduction}, etc. Among these, DTMC model having the properties of both simplicity and effectiveness has a widely application in realistic programs\cite{arenas2010discrete,lange2010discrete}. Classical DTMC does a great job in prediction when the discrete states are easy to distinguish. However, The uncertainty and vagueness exist in real world inevitably. The application of DTMC model is limited when the states are not discrete or the realistic states are uncertain. For example, the states can not be determined according to the known data or the collected data may not be crisp. Fuzzy mathematics is a great tool to handle with uncertainty\cite{zadeh1965fuzzy,Chou2016A,Jiang2015improved}. To address it, some modified models have been proposed like fuzzy Markov chain\cite{buckley2005fuzzy,avrachenkov2002fuzzy} and fuzzy states based on Markov chain\cite{de1998expected,kleyle1997transition}. These models introduce a certain subordinating degree function between the states distribution and the states description. However, considering the complexity of the realistic program, the certain function may be hard to build. Moreover, the evidential Markov chain\cite{Soubaras2010On,Soubaras2009An} and some generalized model models\cite{Zhou2013A,Deng2015Newborns} are also proposed.

In this paper, a new belief Markov chain is proposed. The new model uses the basic probability assignment(BPA) to describe the uncertainty of states as Dempster-Shafer theory\cite{shafer1976mathematical,yager2008classic} is an efficient tool to deal with the uncertainty\cite{Songyf2016,Yang2016A,boujelben2009building,benavoli2008modelling}. The Dempster-shafer fusion can also be modelled in the Markov fields.\cite{Pieczynski2006Multisensor,Boudaren2016Dempster}. Interval number is a simple and efficient tool to handle the uncertain data\cite{jiang2008nonlinear,dou2016interactive}. Considering the properties of interval number, the BPA is generated based on interval number in our model. Due to it, the uncertain data can be represented in the form of interval number. An application in inventory control shows that our proposed model can represent and handle with uncertainty effectively. The prediction result also agrees with the practical situation, which proves the correctness of our model.

The rest of this paper is organized as follows. The preliminaries of the basic theory employed are briefly presented in Section 2. And the shortcoming of DTMC model is illustrated in Section 3. Then our new belief Markov chain model is shown in Section 4. Section 5 uses a numerical example of inventory prediction to show the efficiency of our model. Finally, the paper is concluded in Section 6.
\section{Preliminaries}
In this section, some preliminaries such as DTMC, Dempster-Shafer theory, Pignistic probability transformation(PPT) and interval number are briefly introduced.
\subsection{Discrete-time Markov chain}
\begin{definition}
let $\left\{ {{X_n}:n > 0} \right\}$ be a random sequence defined in the probability space $\left( {\Omega ,F,P} \right)$. $P$ represents probability measure which is a function from set $F$ to filed of real number $R$. Every event in $F$ is given a probability value ranged from 0 to 1 by the function $P$. For arbitrary $n \in {N^ + }$ and states ${i_1},{i_2}, \ldots $, when $P\left\{ {{X_n} = {i_n},{X_{n - 1}} = {i_{n - 1}}, \ldots ,{X_1} = {i_1}} \right\} > 0$ if satisfying
\begin{equation}\label{Eq.1}
P\left\{ {{X_{n + 1}} = {i_{n + 1}}|{X_n} = {i_n}, \ldots ,{X_1} = {i_1}} \right\} = P\left\{ {{X_{n + 1}} = {i_{n + 1}}|{X_n} = {i_n}} \right\}
\end{equation}
then the random sequence $\left\{ {{X_n}:n > 0} \right\}$ is called the Markov chain. Eq.(\ref{Eq.1}) is called the Markov property.
\end{definition}

\begin{definition}
Markov chain $\left\{ {{X_n}:n > 0} \right\}$ is homogeneous if meeting the following condition:
Given arbitrary $m$, $n$ and states $i$, $j$, meeting $P\left\{ {{X_n} = i} \right\} > 0$ and $P\left\{ {{X_m} = i} \right\} > 0$,
\begin{equation}
P\left\{ {{X_{n + 1}} = j|{X_n} = i} \right\} = P\left\{ {{X_{m + 1}} = j|{X_m} = i} \right\}
\end{equation}
\end{definition}

\begin{definition}
For homogeneous Markov chain, the following conditional probability
\begin{equation}
{P_{ij}}\left( {m,m + n} \right) = P\left\{ {{X_{m + n}} = j|{X_m} = i} \right\}
\end{equation}
is called the transition probability from the condition that $m$ moment Markov chain is in state $i$ to the condition that $m+n$ moment Markov chain is in state $j$. And the matrix composed of transferring probability is called transferring probability matrix.
\end{definition}

The following is some important properties of Markov chain.
let $E$ denote state space, $P_{ij}^{(k)}$ denote the probability of state $i$ transferring to state $j$ through $k$ steps. The $k$ transferring probability of homogeneous Markov chain has the following properties:
\begin{equation}
\begin{array}{*{20}{c}}
{P_{ij}^{(k)} \ge 0,}&{\forall i,j \in E,}&{k \ge 0}
\end{array}
\end{equation}
\begin{equation}
\begin{array}{*{20}{c}}
{\sum\limits_{j \in E} {P_{ij}^{(k)} = 1} ,{\kern 1pt} }&{\forall i \in E,}&{k \ge 0}
\end{array}
\end{equation}
\begin{equation}
\begin{array}{*{20}{c}}
{P_{ij}^{(m + k)} = \sum\limits_{r \in E} {P_{ir}^{(m)}P_{rj}^{(k)}} ,{\kern 1pt} }&{\forall i,j \in E,}&{m,k \ge 0}
\end{array}
\end{equation}
\subsection{Dempster-Shafer evidence theory}
Tough evidence theory has some open issues, such as conflict management \cite{deng2015Generalized,yu2015improved}, dependent evidence combination \cite{su2015Combining} and determination of basic probability assignment\cite{Jiang2016A}, it has a wide applications like fault diagnosis\cite{jiang2016AN}, supplier chain management\cite{XinyangDAHP2014156,Deng2011A}, decision making\cite{fu2015group,kang2012evidential,Deng2014D,Li2016The} and risk evaluation which matters a lot in reality\cite{Guo2016A,kabir2015evaluating,feng2015managing}, due to its efficiency to model and fuse uncertain information. The following is some basic concepts of D-S evidence theory.
\begin{definition}
Let $U$ denote a finite set composed of the whole possible value of the random variable $X$. The elements of set $U$ are mutually exclusive and $U$ is called the frame of discernment. Let ${2^U}$ denote the power set of $U$ whose each element corresponds to a subset of the value of $X$.
\end{definition}
\begin{definition}
Let $U$ denote the frame of discernment. Given an arbitrary proposition(subset) $A$ of $U$, a mass function called the basic probability assignment(BPA) is a mapping
$m:{2^U} \to \left[ {0,1} \right]$, satisfying the following conditions:
\begin{equation}
\sum\limits_{A \in {2^U}} {m\left( A \right)}  = 1
\end{equation}
and
\begin{equation}
m\left( \emptyset  \right) = 0{\kern 1pt} {\kern 1pt}
\end{equation}
where ${m\left( A \right)}$ reflects the evidence's supporting degree to the proposition $A$. $A$ is called the focal element if satisfying
$m\left( A \right) > 0$
\end{definition}
\subsection{Pignistic probability transformation}
Since the evidence theory assigns the probability to all the subsets of the frame of discernment, the BPA usually relates to probability assignment of a multi-element subset rather than a singleton subset. It reflects the uncertainty of the real world. However, the decisions are uneasy to make by using BPA directly. The BPA is usually converted to probability and then the decision will be make based on the probability. Pignistic probability transformation(PPT) is a classical method to achieve it by averaging the BPA of a multi-element set into singleten sets.
\begin{definition}
Let $U$ denote the frame of discernment and $m$ be a BPA on $U$. A pignistic probability transformation function
${\rm{Bet}}{\kern 1pt} {\kern 1pt} {\kern 1pt} {\rm{U}} \to \left[ {0,1} \right]$ is defined as:
\begin{equation}
\label{PPT}
B{\rm{et}}{\kern 1pt} {\kern 1pt} {\rm{P}}\left( x \right) = \sum\limits_{x \subseteq A,A \in U} {\frac{{m\left( A \right)}}{{\left| A \right|}}}
\end{equation}
where $\left| A \right|$ is the cardinality of proposition $A$.
\end{definition}
\subsection{Interval number}
\begin{definition}
An interval number
${\tilde a}$ is defined as
$\tilde a = \left[ {{a^ - },{a^ + }} \right] = \left\{ {x|{a^ - } \le x \le {a^ + }} \right\}$
where ${a^ - }$ and ${a^ + }$ are the lower limiting value and upper limiting value apparently while
$x \in \left[ {0,1} \right]$. Especially, interval number ${a^ - }$ degenerates into a real number when ${a^ - } = {a^ + }$.
\end{definition}
\begin{definition}
Let $A = \left[ {{a_1},{a_2}} \right]$ and $B = \left[ {{b_1},{b_2}} \right]$ be two interval numbers. The square of the distance between two interval numbers
${{D^2}\left( {A,B} \right)}$ is calculated by \cite{tran2002comparison}:
\begin{equation}\label{internal distance}
\begin{array}{*{20}{l}}
{{D^2}\left( {A,B} \right){\rm{ = }}\int_{ - \frac{1}{2}}^{\frac{1}{2}} {{{\left\{ {\left[ {\frac{{\left( {{{\rm{a}}_1} + {a_2}} \right)}}{2} + x\left( {{a_2} - {a_1}} \right)} \right] - \left[ {\frac{{\left( {{b_1} + {b_2}} \right)}}{2} + x\left( {{b_2} - {b_1}} \right)} \right]} \right\}}^2}} dx}\\
{{\kern 1pt} {\kern 1pt} {\kern 1pt} {\kern 1pt} {\kern 1pt} {\kern 1pt} {\kern 1pt} {\kern 1pt} {\kern 1pt} {\kern 1pt} {\kern 1pt} {\kern 1pt} {\kern 1pt} {\kern 1pt} {\kern 1pt} {\kern 1pt} {\kern 1pt} {\kern 1pt} {\kern 4pt} {\kern 1pt} {\kern 1pt} {\kern 1pt} {\kern 1pt} {\kern 1pt} {\kern 1pt} {\kern 1pt} {\kern 1pt} {\kern 1pt} {\kern 1pt} {\kern 1pt} {\kern 1pt} {\kern 1pt} {\kern 1pt} {\kern 1pt} {\kern 1pt} {\kern 1pt} {\kern 1pt} {\kern 1pt} {\kern 1pt} {\kern 1pt} {\kern 1pt} {\kern 1pt} {\kern 1pt} {\kern 1pt} {\kern 1pt}  = {{\left[ {\frac{{\left( {{a_1} + {a_2}} \right)}}{2} - \frac{{\left( {{b_1} + {b_2}} \right)}}{2}} \right]}^2} + \frac{{{{\left[ {\left( {{a_2} - {a_1}} \right) + \left( {{b_2} - {b_1}} \right)} \right]}^2}}}{{12}}}
\end{array}
\end{equation}
\end{definition}

\section{The shortcoming of DTMC model}\label{classical application and shortcoming}
In this section, a realistic example of inventory anticipation will show the detailed application of DTMC model and its main shortcoming.
\begin{example}\label{classical example}
Table \ref{crisp inventory} is one company's statistics of the inventory demand for Product E15 in 20 consecutive periods. Relying on these data, the number 21 periods' inventory can be anticipated by applying the traditional DTMC model.
\begin{table}[!h]
\footnotesize
\centering
\caption{Inventory demand for Product E15 in 20 consecutive periods}
\label{crisp inventory}
\begin{tabular*}{\columnwidth}{lllllllllllllllllllllllll}
\toprule
\multicolumn{2}{l}{Time(period)}       &\multicolumn{1}{c}{1}  &\multicolumn{1}{c}{2}  &\multicolumn{1}{c}{3}  &\multicolumn{1}{c}{4}  &\multicolumn{1}{c}{5}  &\multicolumn{1}{c}{6}  &\multicolumn{1}{c}{7}  &\multicolumn{1}{c}{8}  &\multicolumn{1}{c}{9}  &\multicolumn{1}{c}{10}  \\
\multicolumn{2}{l}{Inventory(package)} &143&152&161&139&137&174&142&141&162&180  \\
\midrule
\multicolumn{2}{l}{Time(period)}       &\multicolumn{1}{c}{11}  &\multicolumn{1}{c}{12}  &\multicolumn{1}{c}{13}  &\multicolumn{1}{c}{14}  &\multicolumn{1}{c}{15}  &\multicolumn{1}{c}{16}  &\multicolumn{1}{c}{17}  &\multicolumn{1}{c}{18}  &\multicolumn{1}{c}{19}  &\multicolumn{1}{c}{20}  \\
\multicolumn{2}{l}{Inventory(package)} &164&171  &206  &193  &207  &218  &229  &225  &204  &200  \\
\bottomrule
\end{tabular*}
\end{table}
\end{example}
Generally, the most significant step of prediction is to build a transition probability matrix. Then according to the original states probability assignment, the next period's probability of transferring to each state can be calculated. A larger transferring probability means it is more likely to be in the state. Consequently, the state of next period can be predicted.

First of all, the states space of DTMC model needs to be determined. Based on the data in Table \ref{crisp inventory}, the inventory demand of one period ranges from 137 to 229. If every integer in the interval is deemed as one state, the number of states will be overmuch and the state transferring condition will be hard to count. Considering the number of states should be rational and convenient for building transition probability matrix, the states need to be classified and the classified states are show in Table \ref{The results of classified states}.
\begin{table}[!h]
{\footnotesize
\caption{The results of classified states.}\label{The results of classified states}
\begin{tabular*}{\columnwidth}{@{\extracolsep{\fill}}@{~~}ccccc@{~~}}
\toprule
       Inventory(package)          &[100,150) &[150,200) &[200,250)                             \\
\midrule
       States                      & Low(L) &Medium(M) &High(H)   \\
       Serial number               &1&2&3\\
\bottomrule
\end{tabular*}
}
\end{table}

Let ${N_{ij}}$ denote the times of transferring from the state $i$ to state $j$. Then the following data are obtained.
\[\begin{array}{*{20}{c}}
{{N_{11}} = 2,}&{{N_{12}} = 3,}&{{N_{13}} = 0;}
\end{array}\]
\[\begin{array}{*{20}{c}}
{{N_{21}} = 2,}&{{N_{22}} = 4,}&{{N_{23}} = 2;}
\end{array}\]
\[\begin{array}{*{20}{c}}
{{N_{31}} = 0,}&{{N_{32}} = 2,}&{{N_{33}} = 4.}
\end{array}\]
Let $E$ denote the whole state space. Based on the following equation
\begin{equation}
{P_{ij}} = {N_{ij}}/\sum\nolimits_{j \in E} {{N_{ij}}}
\end{equation}
the transfer probability matrix can be obtained as:
\[P = \left[ {\begin{array}{*{20}{c}}
{0.400}&{0.600}&{0.000}\\
{0.250}&{0.500}&{0.250}\\
{0.000}&{0.333}&{0.667}
\end{array}} \right]\]
The inventory of 20th period is 200 packages, belonging to the state medium. Based on the obtained transfer probability matrix, the state probability assignment of the next period is:
$\left( {L,M,H} \right) = \left( {0.250,0.500,0.250} \right)$ Thus, the 21st period's state is most likely to be medium with the probability of 0.500.

However, a serious shortcoming exists in this model. If the inventory of the final period changes from 200 to 201 and the rest data keep the same, then the new transition probability matrix is obtained as:
\[P = \left[ {\begin{array}{*{20}{c}}
{0.400}&{0.600}&{0.000}\\
{0.250}&{0.500}&{0.250}\\
{0.000}&{0.167}&{0.833}
\end{array}} \right]\]
Because 201 belongs to the state high, the state probability assignment of 21st period turns into:
$\left( {L,M,H} \right) = \left( {0.000,0.167,0.8333} \right)$.
And now the state high becomes the most likely one with a probability of 0.833, which is much lager than 0.500.
The tiny change in the final period leads to a drastic change of the prediction, which is obviously irrational. The prediction result should be close whatever the final inventory is 200 or 201. The reason causing this situation is that the states are classified by a too crisp critical region. Each value belongs to a certain state with a probability of 1 completely. Thus, the prediction result may change drastically in the critical region.

In realistic programs, the value should not correspond to certain state completely. For example, 200 may belong to the state medium with a probability of 0.4 while to the state high with a probability of 0.6. Or the value can not be classified into certain, like it may belong to both medium and high, but the probability of belong to medium or high is uncertain. Besides, the data of the previous periods or the current period may be uncertain. In that case, the states may be hard to classified and the transferring probability matrix is difficult to build. To address these problems, a new belief Markov model is proposed.

\section{The new belief Markov chain model}\label{Proposed interval belief Markov chain model}
The integrated process of building the new belief Markov model is as following:

\textbf{Step 1.} Determine the state space based on the previous data. Make sure the number of states is rational and all the states form the frame of discernment $U$.

\textbf{Step 2.} Calculate the BPA of the whole periods on the power set of the discernment ${2^U}$ based on the distance between interval numbers.

\textbf{Step 3.} Calculate the single-step transferring belief assignment ${P_{ij}}$, $i,j \in {2^U}$ based on the obtained BPA, . And then the transition belief matrix $\left[ {{P_{ij}}} \right]$ is built. ${P_{ij}}$ represents the belief assignment of transferring from proposition $i$ to proposition $j$:
\begin{equation}\label{P_ij}
{P_{ij}} = \begin{array}{*{20}{c}}
{\frac{{\sum\limits_{t = 1}^{n - 1} {\left( {m{{\left( i \right)}_t} \cdot m{{\left( j \right)}_{t + 1}}} \right)} }}{{\sum\limits_{k \in {2^U}} {\sum\limits_{t = 1}^{n - 1} {\left( {m{{\left( i \right)}_t} \cdot m{{\left( k \right)}_{t + 1}}} \right)} } }},}&{i,j \in {2^U}}
\end{array}
\end{equation}
where ${m{{\left( i \right)}_t}}$ represents the assigned belief of proposition $i$ in the t moment, $n$ represents the length of the Markov chain.
\begin{note}
In classical DTMC model, the transition takes place between basic states. While in new belief Markov chain model, the transition takes place between propositions. In other words, the probability is replaced with BPA. Hence, the original transition probability matrix is replaced with the transition belief matrix.
\end{note}

\textbf{Step 4.} Let
$m = \begin{array}{*{20}{c}}
{\left[ {m\left( i \right)} \right],}&{i \in {2^U}}
\end{array}$ be the BPA of the final period. Then the assignment of the next period can be obtained by:
\begin{equation}\label{new m'}
m' = m \cdot \left[ {{P_{ij}}} \right]
\end{equation}

\textbf{Step 5.}
Convert the obtained BPA of the next period $m'$ into the states probability assignment
$\begin{array}{*{20}{c}}
{\left[ {{P_{ij}}} \right],}&{i \in {2^U}}
\end{array}$ by using PPT. And the final prediction result is obtained.

In our model, one of the most significant step is to generate BPA. The detailed process of generating BPA is shown in the following.
\begin{figure}[!ht]
\centering
\includegraphics[scale=0.65]{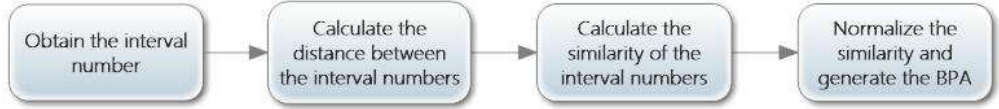}
 \caption{The flow chart of generating BPA}\label{BPAgenerate}
\end{figure}
First of all, the interval BPA need to be obtained. The fuzziness and uncertainty existed in realistic situation can be effectively represented in the form of interval number. And the crisp number can also be deemed as an interval number like 0.5 can be seen as [0.5,0.5].

By using Eq.(\ref{internal distance}), the distance between these interval numbers can be calculated.

Then the similarity of the interval numbers can be obtained based on the distance.
\begin{definition}
Let $A{\rm{ = }}\left[ {{{\rm{a}}_1},{a_2}} \right]$ and $B{\rm{ = }}\left[ {{{\rm{b}}_1},{b_2}} \right]$ be two interval numbers. The similarity of the two interval numbers
$S\left( {A,B} \right)$ is defined as:
\begin{equation}\label{similarity}
S\left( {A,B} \right) = \frac{1}{{1 + {D^2}\left( {A,B} \right)}}
\end{equation}
where $D\left( {A,B} \right)$ is the distance between interval number $A$ and $B$.
\end{definition}
When the interval number $A$ equals to $B$, $S\left( {A,B} \right) = 1$. According to the definition, it is easy to know that the larger the difference between $A$ and $B$ is, the similarity is smaller.

Finally, normalize the obtained similarity and the BPA of interval number is obtained. An example will show the process in the following.
\begin{example}\label{Example to generate BPA}
Let state $A$ range in the interval (0,5] and state $B$ range in the interval (5,10]. Given an interval number $C$=[3,6] and $\alpha$=3, the result of generated BPA is shown as Table \ref{The example BPA result}.
\begin{table}[!h]
{\renewcommand\arraystretch{1.2}
\caption{The result of example \ref{Example to generate BPA}.}\label{The example BPA result}
\begin{tabular*}{\columnwidth}{@{\extracolsep{\fill}}@{~~}llllllll@{~~}}
\toprule
       &States          &Distance &Similarity &BPA                             \\
\midrule
       &\multicolumn{1}{c}{A}   &\multicolumn{1}{c}{${\textstyle{{28} \over 3}}$}  &\multicolumn{1}{c}{${\textstyle{{3} \over 31}}$} &\multicolumn{1}{c}{0.5974}   \\
       &\multicolumn{1}{c}{B}   &\multicolumn{1}{c}{${\textstyle{{43} \over 3}}$}  &\multicolumn{1}{c}{${\textstyle{{3} \over 46}}$} &\multicolumn{1}{c}{0.4026}\\
\bottomrule
\end{tabular*}
}
\end{table}
\end{example}

%
\section{Numerical example}\label{Numerical example}
Inventory control is a common realistic problem\cite{Levi2005Approximation,Ozbay2007Stochastic,Memari2014Production}. In this section, the inventory prediction problem in Section \ref{classical application and shortcoming} will still be taken as an example to show the effectiveness of our model. We assume that some uncertain information exist in the realistic statistics, some data of inventory demand are in the form of interval number. Table \ref{interval inventory} is one company's statistics of the inventory demand for Product E15 in 20 consecutive periods. Following the steps described in Section \ref{Proposed interval belief Markov chain model}, the new belief Markov model will be applied to do the prediction.
\begin{table}[!h]
\footnotesize
\centering
\caption{Inventory demand for Product E16 in 20 consecutive periods}
\label{interval inventory}
\begin{tabular*}{\columnwidth}{lllllllllllllllllllllllll}
\toprule
\multicolumn{2}{c}{Time(period)}       & \multicolumn{1}{c}{1}  & \multicolumn{1}{c}{2}  & \multicolumn{1}{c}{3}  & \multicolumn{1}{c}{4} & \multicolumn{1}{c}{5} & \multicolumn{1}{c}{6}  & \multicolumn{1}{c}{7}  & \multicolumn{1}{c}{8}  & \multicolumn{1}{c}{9}  & \multicolumn{1}{c}{10} \\
\multicolumn{2}{l}{Inventory(package)} & 143          & 152                    & {[}157,162{]}                    & 139                   & 137                   & {[}165,180{]}          & 142                    & 141                    & 162                    & 180          \\
\midrule
\multicolumn{2}{c}{Time(period)}       & \multicolumn{1}{c}{11} & \multicolumn{1}{c}{12} & \multicolumn{1}{c}{13} & 14                    & 15                    & \multicolumn{1}{c}{16} & \multicolumn{1}{c}{17}    & \multicolumn{1}{c}{18} & \multicolumn{1}{c}{19} & \multicolumn{1}{c}{20} \\
\multicolumn{2}{l}{Inventory(package)} & 164          & 171                    & {[}204,209{]}                & 193                   & 207                   & {[}215,220{]}          & 229                        & 225                    & 204                    & 200\\
\bottomrule
\end{tabular*}
\end{table}

\begin{figure}[!ht]\label{zhexiantu}
\centering
\includegraphics[scale=0.65]{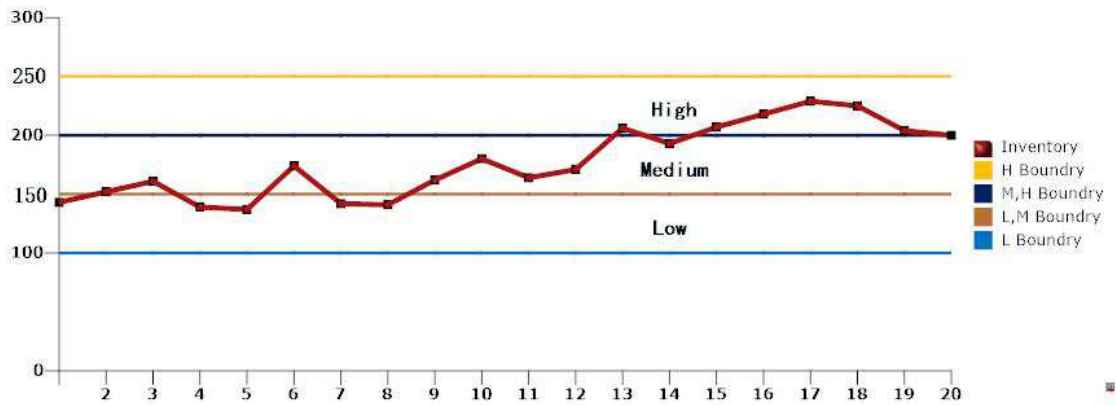}
\caption{Previous inventory demand}
\end{figure}

 As shown in Table \ref{interval inventory}, the inventory of the whole 20 periods ranges in the interval number [137, 229]. All the values can be classified into three basic states: low(L), medium(M) and high(H). They constitute the frame of discernment $U = \left\{ {L,M,H} \right\}$. Then the power set of the frame is consisted of $\left\{ {L} \right\}$, $\left\{ {M} \right\}$, $\left\{ {H} \right\}$, $\left\{ {L,M} \right\}$, $\left\{ {M,H} \right\}$, $\left\{ {L,H} \right\}$, $\left\{ {L,M,H} \right\}$ and the empty set $\emptyset $. To assign the value of inventory into its according proposition, the correspondences between propositions and the value of inventory are determined as Table \ref{Propositions and inventory}. Considering the realistic situation, the assessment of the inventory can not be both low and high, nor do the propositions $\left\{ {L,M,H} \right\}$ or empty set $\emptyset $. So these propositions do not have a according inventory value. After assigning, the distribution of the 20 periods can be showed like Figure \ref{zhexiantu}.
\begin{table}[!h]
\centering
\caption{Correspondences between propositions and inventory}
\label{Propositions and inventory}
\begin{tabular}{ccc}
\toprule
\textbf{Series} &\textbf{Propositions} & \textbf{Inventory} \\
\midrule
1&\{L\}                 & {[}100,150{]}      \\
2&\{L,M\}               & {[}135,165{]}      \\
3&\{M\}                 & {[}150,200{]}      \\
4&\{M,H\}               & {[}185,215{]}      \\
5&\{H\}                 & {[}201,250{]}      \\
-&\{L,H\}               & -                  \\
-&\{L,M,H\}             & -                  \\
-&$\emptyset $          & -                  \\
\bottomrule
\end{tabular}
\end{table}

Generally speaking, given an inventory value, its according proposition is certain. However, in our model, it can be assigned to every proposition with different probabilities. The probabilities of the assignment are unknown for now. The BPA cannot be obtained without a certain probability. Thus the following step is to calculate the probabilities of assignment.

Let us take the inventory of 6th period [165,180] as an example. Based on Eq.(\ref{internal distance}), the distance among the inventory and all valid propositions can be obtained. Then by using Eq.(\ref{similarity}), the similarity among the inventory and all valid propositions can also be obtained. Lastly, the BPA of 6th period can be obtained by normalizing the similarity. The results are shown in Table \ref{The BPA of 6th period}. And the BPA of 6th period is
\[{\rm{m}}\left( {\left\{ L \right\},\left\{ {L,M} \right\},\left\{ M \right\},\left\{ {M,H} \right\},\left\{ H \right\}} \right) = [0.0324,0.1422,0.7033,0.0964,0.0257]\]
\begin{table}[!ht]
\centering
\caption{The BPA of 6th period}
\label{The BPA of 6th period}
\begin{tabular}{cccc}
\toprule
\textbf{Propositions} & \textbf{Distance} & \textbf{Similarity} & \textbf{BPA}    \\
\midrule
\{L\}       & 48.0139  & 4.3359$ \times $e-4 & 0.0324 \\
\{L,M\}     & 22.9129  & 0.0019              & 0.1422 \\
\{M\}       & 10.2632  & 0.0094              & 0.7033 \\
\{M,H\}     & 27.8388  & 0.0013              & 0.0964 \\
\{H\}       & 53.9011  & 3.4408$ \times $e-4 & 0.0257 \\
\bottomrule
\end{tabular}
\end{table}

Repeat the above process for 20 times, the BPA of each period can be obtained in Table \ref{BPA of 20 periods}.
\begin{table}[!ht]
\caption{The BPA of 20 periods}
\label{BPA of 20 periods}
\centering
\noindent
\begin{tabular}{|c||*{5}{c|}}\hline
\backslashbox{Period}{Propositions}
&\{L\}&\{L,M\}&\{M\}
&\{M,H\}&\{H\}\\\hline\hline
1  & 0.1758 & 0.7137  &  0.0709 &   0.0268  &  0.0127 \\\hline
2  & 0.0713 & 0.8047   & 0.0855   & 0.0270   & 0.0115   \\\hline
3  & 0.0694 & 0.6379  &  0.2186  &  0.0540 &   0.0202 \\\hline
4  & 0.2988 & 0.5814   & 0.0747  &  0.0302 &   0.0149  \\\hline
5  & 0.3728 & 0.5072  &  0.0738   & 0.0307   & 0.0155  \\\hline
6  & 0.0324 & 0.1422  &  0.7033  &  0.0964  &  0.0257  \\\hline
7  & 0.2023 & 0.6841 &   0.0724 &   0.0278  &  0.0134  \\\hline
8  & 0.2319 & 0.6518  &  0.0736  &  0.0288 &   0.0139  \\\hline
9  & 0.0750 & 0.5224    &0.2998 &   0.0756  &  0.0272  \\\hline
10 & 0.0375 & 0.1220  &  0.5379  &  0.2501 &   0.0524 \\\hline
11 & 0.0720 & 0.4456  &  0.3636  &  0.0883 &   0.0304 \\\hline
12 & 0.0296 & 0.0981 &   0.6195  &  0.2111&    0.0417  \\\hline
13 & 0.0108 & 0.0224  &  0.0648  &  0.7630 &   0.1390  \\\hline
14 & 0.0183 & 0.0452  &  0.1715  &  0.6959 &   0.0692 \\\hline
15 & 0.0131 & 0.0270  &  0.0753  &  0.7191 &   0.1654 \\\hline
16 & 0.0158 & 0.0295 &   0.0691 &   0.3467 &   0.5389\\\hline
17 & 0.0144 & 0.0249  &  0.0514 &   0.1716  &  0.7377  \\\hline
18 & 0.0141 & 0.0249 &   0.0535 &   0.2025 &   0.7050 \\\hline
19 & 0.0113 & 0.0241 &   0.0712 &   0.7845 &   0.1088 \\\hline
20 & 0.0108 & 0.0240 &   0.0773  &  0.8151  &  0.0728 \\\hline
\end{tabular}
\end{table}

As shown in Table \ref{BPA of 20 periods}, in this program every BPA is nonzero. Thus the Eq.(\ref{P_ij}) can be rewritten as following:
\begin{equation}\label{New P_ij}
{P_{ij}} = \begin{array}{*{20}{c}}
{\frac{{\sum\limits_{t = 1}^{n - 1} {\left( {m{{\left( i \right)}_t} \cdot m{{\left( j \right)}_{t + 1}}} \right)} }}{{\sum\limits_{k \in {2^U}} {\sum\limits_{t = 1}^{n - 1} {\left( {m{{\left( i \right)}_t} \cdot m{{\left( k \right)}_{t + 1}}} \right)} } }} = \frac{{\sum\limits_{t = 1}^{n - 1} {\left( {m{{\left( i \right)}_t} \cdot m{{\left( j \right)}_{t + 1}}} \right)} }}{{\sum\limits_{t = 1}^{n - 1} {m{{\left( i \right)}_t}} }},}&{i,j \in {2^U}}
\end{array}
\end{equation}
By using Eq.(\ref{New P_ij}), the belief assignment of transferring from one proposition to another proposition can be calculated. For example, the belief assignment of transferring from proposition $\left\{ {L} \right\}$ to proposition $\left\{ {L,M} \right\}$ is calculated as:
\begin{equation}
{P_{12}} = \frac{{\sum\limits_{t = 1}^{n - 1} {\left( {m{{\left( {\left\{ L \right\}} \right)}_t} \cdot m{{\left( {\left\{ {L,M} \right\}} \right)}_{t + 1}}} \right)} }}{{\sum\limits_{t = 1}^{n - 1} {m{{\left( {\left\{ L \right\}} \right)}_t}} }} = 0.4208
\end{equation}
All the obtained transferring belief assignments constitute a 5$\times$5 transition belief matrix $\left[ {{P_{ij}}} \right]$ as following:
\[\left[ {\begin{array}{*{20}{c}}
{{\rm{0}}{\rm{.1369}}}&{{\rm{0}}{\rm{.4208}}}&{{\rm{0}}{\rm{.2914}}}&{{\rm{0}}{\rm{.1081}}}&{{\rm{0}}{\rm{.0427}}}\\
{{\rm{0}}{\rm{.1329}}}&{{\rm{0}}{\rm{.4712}}}&{{\rm{0}}{\rm{.2559}}}&{{\rm{0}}{\rm{.1046}}}&{{\rm{0}}{\rm{.0353}}}\\
{{\rm{0}}{\rm{.0923}}}&{{\rm{0}}{\rm{.3188}}}&{{\rm{0}}{\rm{.2250}}}&{{\rm{0}}{\rm{.2769}}}&{{\rm{0}}{\rm{.0870}}}\\
{{\rm{0}}{\rm{.0299}}}&{{\rm{0}}{\rm{.0938}}}&{{\rm{0}}{\rm{.1271}}}&{{\rm{0}}{\rm{.5199}}}&{{\rm{0}}{\rm{.2292}}}\\
{{\rm{0}}{\rm{.0228}}}&{{\rm{0}}{\rm{.0611}}}&{{\rm{0}}{\rm{.0895}}}&{{\rm{0}}{\rm{.4165}}}&{{\rm{0}}{\rm{.4100}}}
\end{array}} \right]\]

The next step is to predict the inventory of the 21st period based on the BPA of the 20th period and the matrix $\left[ {{P_{ij}}} \right]$.
As Table \ref{BPA of 20 periods} shows, the BPA of the 20th is $m\left( {20} \right) = \left( {0.0228,0.0611,0.0895,0.4165,0.4100} \right)$.
By using the Eq.(\ref{new m'}) the BPA of the 21st is calculated as:
\begin{equation}
m\left( {21} \right) = m\left( {20} \right) \cdot \left[ {{P_{ij}}} \right] = ({\rm{0}}{\rm{.0379,0}}{\rm{.1214,0}}{\rm{.1368,0}}{\rm{.4792,0}}{\rm{.2247}})
\end{equation}

Finally, use Pignistic probability transformation(Eq.(\ref{PPT})) to transfer the obtained BPA to the probabilities of basic states.
\[Bet{\kern 1pt} {\kern 1pt} {\kern 1pt} {\kern 1pt} P\left( L \right){\rm{ = }}0.0379 + \frac{{0.1214}}{2} = 0.0986\]
\[Bet{\kern 1pt} {\kern 1pt} {\kern 1pt} {\kern 1pt} P\left( M \right){\rm{ = }}\frac{{0.1214}}{2} + 0.1368 + \frac{{0.4792}}{2} = 0.4371\]
\[Bet{\kern 1pt} {\kern 1pt} {\kern 1pt} {\kern 1pt} P\left( H \right){\rm{ = }}\frac{{0.4792}}{2} + 0.2247 = 0.4643\]
The result is $\left( {L,M,H} \right) = \left( {0.0986,0.4371,0.4643} \right)$. Thus, the inventory of 21st period is most likely to be in stage high. The practical inventory demand for 21st is 223, which is in the state high. Thus, the prediction result confirms with the reality and is rational.

When the inventory of the 20th inventory changes, the result of prediction also change fluently, which verifies the effectiveness and rationality of our model.
\begin{figure}[!ht]\label{The prediction of the 21th period}
\centering
\includegraphics[scale=0.65]{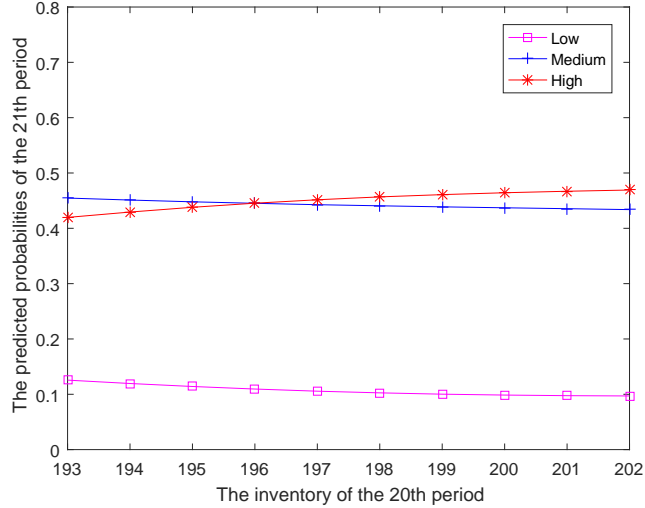}
\caption{The prediction of the 21th period}
\end{figure}
As Figure 3 shows, when the inventory of the 20th period is less than 195, the prediction result is medium. On the contrary, the result is high when the inventory of the 20th is more than 196. And the line changes fluently. It reveals that the shortcoming of the classical Markov model mentioned in Section \ref{classical application and shortcoming} can be effectively solved in our model. In addition, the comparison between the prediction and the real situation is shown in Table \ref{Comparison between prediction and practical situation}. The practical inventory of 20th is 200, thus the prediction result is in accord with the practical situation which prove the effectiveness of our model.
\begin{table}[!h]
\centering
\footnotesize
\caption{Comparison between prediction and practical situation}
\label{Comparison between prediction and practical situation}
\begin{tabular}{|c|c|c|c|c|c|}
\hline
Inventory                    & 193       & 194       & 195       & 196       & 197       \\ \hline
Prediction result(Probability) & M(0.4577) & M(0.4511) & M(0.4479) & H(0.4454) & H(0.4516) \\ \hline
Practical situation          & \multicolumn{5}{c|}{H}                                    \\ \hline
Inventory                    & 198       & 199       & 200       & 201       & 202       \\ \hline
Prediction result(Probability) & H(0.4568) & H(0.4609) & H(0.4643) & H(0.4670) & H(0.4692) \\ \hline
Practical situation          & \multicolumn{5}{c|}{H}                                    \\ \hline
\end{tabular}
\end{table}

In the following, the comparison between our model and classical DTMC model is made. Compared with the classical model, the most significant difference of our model is that the state is distributed into each possible state with a probability. Figure 4 illustrates the methods of state distribution in different models.
\begin{figure}[!ht]\label{state comparison}
\centering
\includegraphics[scale=0.8]{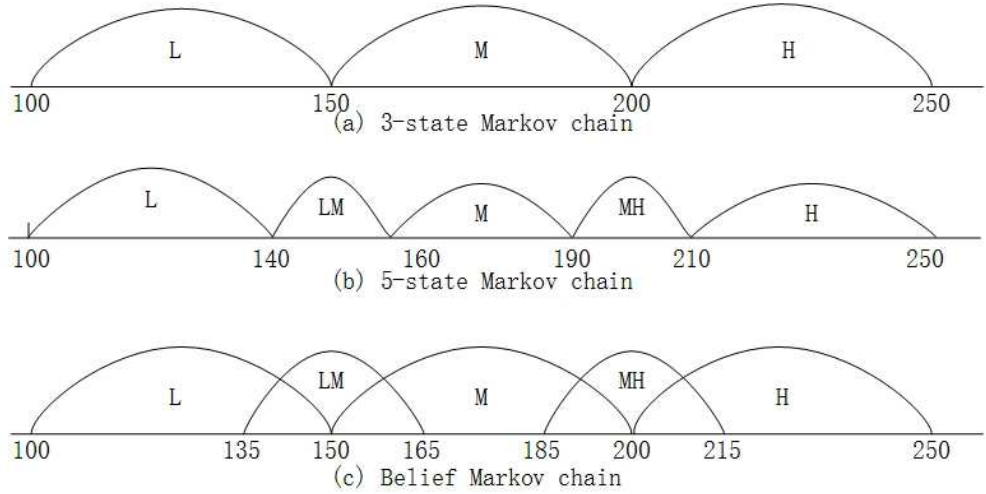}
\caption{Comparison of state distribution}
\end{figure}
The three different models are applied to predict the 21st inventory demand when the data of 20th fluctuates.
\begin{figure}[!ht]\label{modelcompare}
\centering
\includegraphics[scale=1]{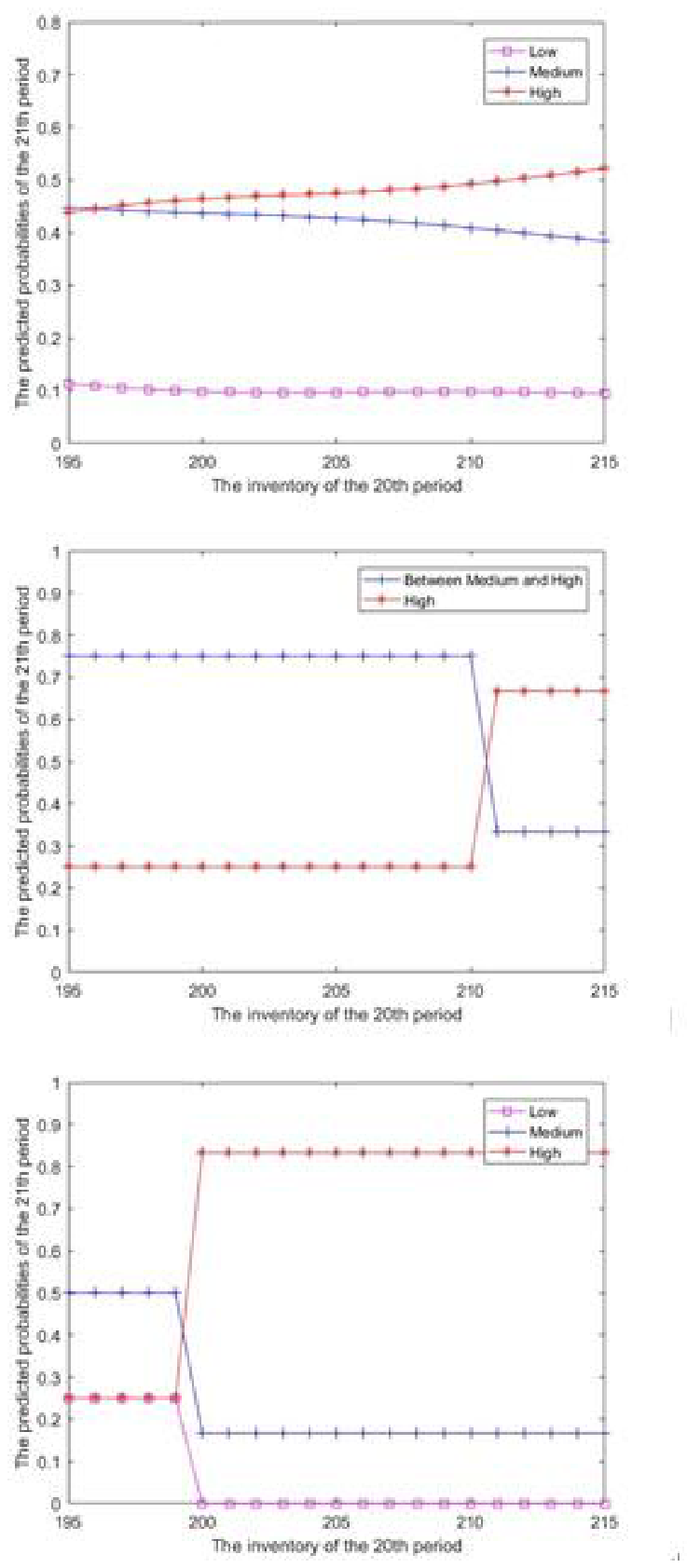}
\caption{Comparison of model prediction}
\end{figure}
As Figure \ref{modelcompare} shows, along the 20th inventory demand changes, the sudden change will exist in prediction of classical Markov chain model.

As mentioned above, the classical DTMC model can not handle the uncertainty, especially a little change of data may lead to a drastic prediction result, which is obviously irrational. The increase in the state number may decrease the trouble caused by uncertainty of states in some degree. However, a larger data set will be needed to assure it and the problem of drastic change in prediction is still unavoidable. By contrast, our new belief Markov chain model can handle these shortcomings effectively. Moreover, the uncertain data like interval number can also be well handled, which proves our model's ability of handling uncertain information.

\section{Conclusion}\label{Conclusion}
In this paper, a new belief Markov chain model is proposed. The shortcomings of classical DTMC model are successfully overcame in this model. The main advantages of the new model are as following:

1. Stability. The sudden change of prediction result is avoided, namely the prediction result will change fluently along the slight change of data.

2. The ability of handling uncertain information. Either the uncertainty of state or data can be effectively handled in our model.

3. Flexibility. The different state distribution schemes and basic probability assignment distribution functions can lead to different results. The setting can be adjusted according to the realistic situation which reveals the flexibility of our model.

A numerical example of inventory prediction is illustrated in the paper to show the application of our model. And the prediction result and comparison prove the effectiveness and rationality of our model.

\section*{Acknowledgement}
The work is partially supported by National Natural Science Foundation of China (Grant No. 61671384), Natural Science Basic Research Plan in Shaanxi Province of China (Program No. 2016JM6018), the Fund of SAST (Program No. SAST2016083), the Seed Foundation of Innovation and Creation for Graduate Students in Northwestern Polytechnical University (Program No. Z2016122).

\bibliographystyle{model2-names}
\bibliography{myreference}
\clearpage


\end{document}